\documentclass{article}

\usepackage[preprint]{corl_2026}

\usepackage{booktabs}
\usepackage[table]{xcolor}
\definecolor{mydarkblue}{rgb}{0,0.08,0.45}
\usepackage{soul, color, xcolor}
\usepackage{graphicx}
\usepackage{wrapfig}
\usepackage{multirow}
\usepackage{array} 
\usepackage{makecell}
\usepackage{utfsym}
\usepackage{fontawesome}
\usepackage{cancel}
\usepackage{enumitem}
\usepackage{amsmath}
\usepackage{bm}
\usepackage{array}
\usepackage{marvosym}
\usepackage{tabularx}

\usepackage{times}
\usepackage{graphicx}
\usepackage{calc}
\usepackage[notransparent]{svg}
\usepackage{amssymb}
\usepackage{amsmath}
\usepackage{xcolor}
\definecolor{cvprblue}{rgb}{0.21,0.49,0.74}

\usepackage{booktabs} \usepackage{multirow} \usepackage{multicol} \usepackage[flushleft]{threeparttable}
\usepackage{colortbl} \usepackage{diagbox}  \usepackage{siunitx}

 \usepackage{xspace} \usepackage[subrefformat=parens,labelformat=parens,font=small]{subcaption} \usepackage{wrapfig}

\usepackage{algorithm}
\usepackage{listings}
\usepackage{wrapfig}
\usepackage{pgfplots}
\usepackage{comicneue}
\usetikzlibrary{patterns}
\pgfplotsset{compat=1.18}
\newcommand{\boldparagraph}[1]{\vspace{0.1cm}\noindent{\bf #1}}

\usepackage[T1]{fontenc} 
\usepackage[table]{xcolor}

\title{Relational Spatial Alignment for Vision-Language-Action Models}

\author{\begin{tabular}
{c@{\hspace{0.6em}}c@{\hspace{0.6em}}c@{\hspace{0.6em}}c@{\hspace{0.6em}}c@{\hspace{0.6em}}c}
Chaoran Zhu$^{1}$ &
Hengyi Wang$^{2}$ &
Chenglin Cui$^{1}$ &
Yazhe Wan$^{1}$ &
Taein Kwon$^{3*}$ &
Changjae Oh$^{1*}$
\end{tabular}
\\[1.2ex]
$^{1}$Queen Mary University of London, United Kingdom \\
$^{2}$University College London, United Kingdom \qquad
$^{3}$VGG, University of Oxford, United Kingdom
\\[1.2ex]
\texttt{\href{https://qm-ipalab.github.io/RIGID}
{qm-ipalab.github.io/RIGID}}
\vspace{-20pt}
}

\hypersetup{
    pdftitle={Relational Spatial Alignment for Vision-Language-Action Models},
    pdfauthor={Chaoran Zhu, Hengyi Wang, Chenglin Cui, Yazhe Wan, Taein Kwon, Changjae Oh}
}

\begin{document}
{\setlength{\tabcolsep}{0pt}\maketitle}

\renewcommand\thefootnote{}
\footnotetext{*Equal supervision and corresponding authors.}
\begin{abstract}
Spatial understanding is essential for robots to perceive their surrounding environments and execute actions. Existing approaches typically encode 3D awareness into vision-language-action (VLA) models by directly aligning their representations with pre-trained 3D foundation models. However, we observe that such direct geometric feature alignment essentially exploits 3D foundation priors as a form of low-rank regularization in the feature space, resulting in severe rank collapse of the pre-trained VLA representations. In contrast, we shift the focus from enforcing point-wise equality to aligning pair-wise relationships. We propose a representation-alignment method that enables the injection of 3D spatial awareness without compromising the model's original expressivity.
Extensive experiments on public benchmarks, our new spatial reasoning benchmark, and real-world manipulation scenarios demonstrate that our approach effectively incorporates 3D spatial representations into VLA reasoning, achieving robust performance and enhanced spatial awareness across diverse manipulation tasks.

\end{abstract} 

\section{Introduction}

Spatial understanding is fundamental to robotic manipulation, enabling robots to interpret their environment from multi-view RGB observations, estimate distances, localize objects, and infer 3D structure either explicitly or implicitly. Traditional approaches~\cite{dexnet2, dope} typically follow a \textit{reconstruction-before-action} paradigm, in which an explicit 3D scene is first reconstructed and then used for downstream tasks such as path planning and control~\cite{octomap,voxblox}.

Inspired by the success of end-to-end learning in natural language processing~\cite{gpt4} and autonomous driving~\cite{uniad,end2endsd}, recent attention has shifted toward Vision-Language-Action (VLA) models, a unified end-to-end learning framework for embodied AI. To incorporate spatial awareness, some prior methods~\cite{perceiver,3dvla,voxposer,pointvla} follow the reconstruction-before-action paradigm by leveraging explicit spatial representations, such as depth maps or point clouds, as additional inputs to VLA models. However, such approaches often depend on pre-scanned environments or high-quality sensor observations, both of which can be unreliable in real-world scenarios due to sensor noise, reflective surfaces, and occlusions. Consequently, such methods remain limited in constrained environments and struggle to generalize to open-world problems.

To avoid reliance on explicit 3D representations, recent works~\cite{spatial-forcing, rocket} shift the focus to encode implicit 3D geometric structure within the latent space. Building on recent advances in 3D foundation models~\cite{dust3r,spann3r,vggt}, this approach transfers geometric knowledge into VLAs through feature-level distillation. While these approaches bypass the need for explicit 3D representations, the underlying mechanism of this latent geometric transfer remains underexplored. More importantly, it raises a fundamental question: \textit{Does injecting spatial awareness into the VLA with direct feature alignment compromise its original semantic-level representational capabilities?}

\begin{figure}[t]
    \centering
    
\definecolor{nblue}{HTML}{1F77B4}
    \definecolor{nred}{HTML}{D62728}
    \definecolor{ngreen}{HTML}{2CA02C}
    \definecolor{norange}{HTML}{BF5700}

\pgfplotsset{
            neurips_style/.style={
                width=\textwidth,
                height=5.9cm, 
                label style={font=\footnotesize },
                tick label style={font=\scriptsize  },
                legend style={
                    font=\scriptsize, 
                    draw=none, 
                    fill=white, 
                    fill opacity=0.8, 
                    cells={anchor=west}
                },
ymajorgrids=true,
                grid style={dashed, line width=.2pt, draw=gray!40},
                axis line style={line width=.6pt, black!80},
                tick style={line width=.6pt, black!80},
legend image code/.code={
                    \draw[#1] (0pt,-0.6ex) rectangle (0.18cm,0.8ex);
                }
            }
        }
\begin{subfigure}[t]{0.49\textwidth}
        \vspace{0pt} \centering
        \includegraphics[width=\linewidth]{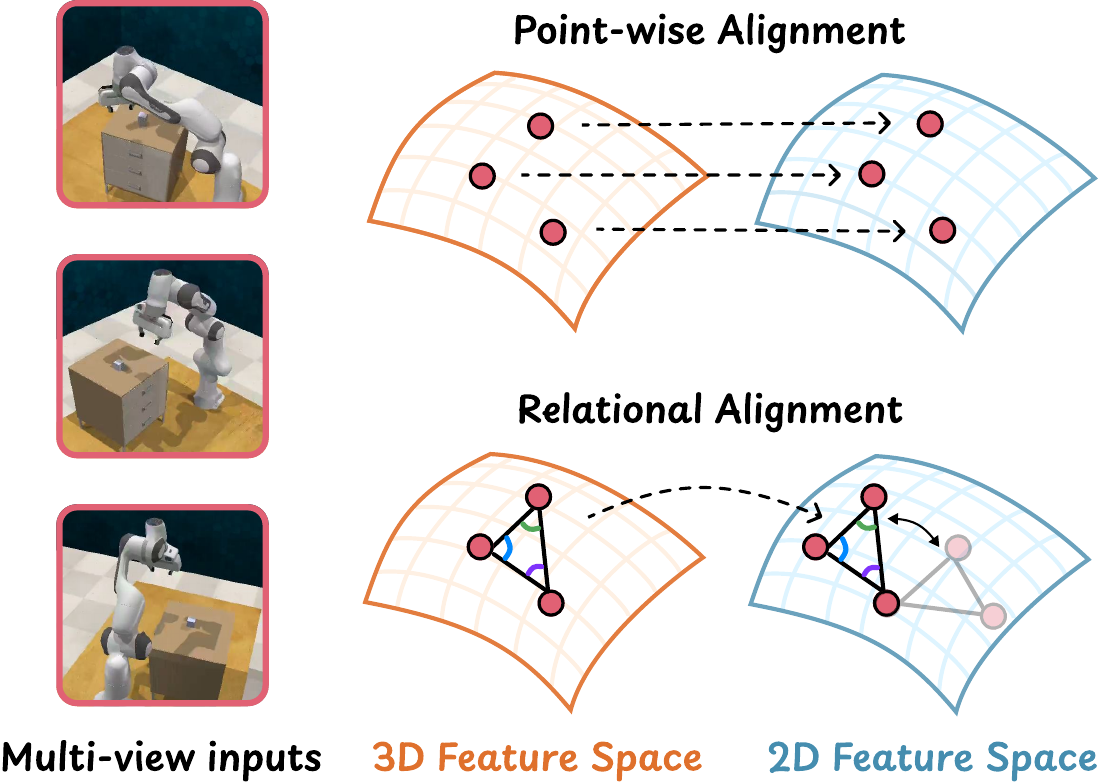}\\[1.5ex]
        \textbf{\footnotesize (a) Alignment Method Comparison}\\
    \end{subfigure}
    \hfill \begin{subfigure}[t]{0.49\textwidth}
        \vspace{0pt} \centering  
        \begin{tikzpicture} 
            \pgfplotsset{
                rank_style/.style={
                    neurips_style,
                    enlarge x limits=0.2,
                    symbolic x coords={3D Model, Abs. Align, Origin, Ours},
                    xtick=data,
                    xtick style={draw=none},   tick label style={font=\comicneue\bfseries\scriptsize}, 
legend style={font=\comicneue\bfseries\scriptsize},     
                    every node near coord/.append style={
font=\comicneue\bfseries\tiny,  
                        /pgf/number format/fixed, 
                        /pgf/number format/precision=1
                    },
                    ybar=0pt,
                    bar width=10pt
                }
            }
\begin{axis}[
                    rank_style,
                    bar shift=-5.5pt, 
                    axis y line*=left,
                    ymin=0, ymax=15,
                    y axis line style={draw=nblue, thick},
                    ytick style={color=nblue, thick},
                    yticklabel style={color=nblue},
                    every node near coord/.append style={xshift=-2pt},  legend pos=north west,
                ]
                \addplot[
                    fill=nblue!20,            
                    draw=nblue, 
                    thick,
                    pattern=north east lines, 
                    pattern color=nblue!80    
                ] coordinates {
                    (Origin, 10.5) (Abs. Align, 6.0) (3D Model, 1.2) (Ours, 11.1)
                };
                \addlegendentry{Stable Rank}
    
\addplot[
                    fill=norange!20, 
                    draw=norange, 
                    thick, 
                    pattern=north east lines, 
                    pattern color=norange!80
                ] coordinates {(Origin, -1)};
                \addlegendentry{Effective Rank}
    
\draw[nred, -stealth, thick] 
                    ([xshift=-3.5pt]axis cs:Origin, 10.8) 
                    to [bend right=30] 
                    node [above=10pt, pos=0.8, sloped, font=\comicneue\bfseries\small, text=nred] {collapse} 
                    ([xshift=-5.5pt, yshift=5pt]axis cs:Abs. Align, 6.5);
            \end{axis}

\begin{axis}[
                rank_style,
                bar shift=5.5pt,
                axis y line*=right,
                axis x line=none,
                ymin=0, ymax=400, 
                y axis line style={draw=norange, thick},
                ytick style={color=norange, thick},
                yticklabel style={color=norange},
                every node near coord/.append style={xshift=4pt},   legend style={draw=none, fill=none},
            ]
                \addplot[
                    fill=norange!20, 
                    draw=norange, 
                    thick,
                    pattern=north east lines, 
                    pattern color=norange!80
                ] coordinates {
                    (Origin, 300.5) (Abs. Align, 249.7) (3D Model, 21.5) (Ours, 306.3)
                };
            \end{axis}
        \end{tikzpicture}
        \\[1.0ex]
        \vspace{-5pt}
\textbf{\footnotesize (b) Rank Comparison}
    \end{subfigure}

    \caption{\textbf{Spectral and rank analysis of different alignment methods.} (a) Traditional 3D knowledge distillation aligns the point-wise magnitudes (i.e, vector norm) or angles (i.e, cos-similarity) across feature spaces, which often leads to rank collapse (b). In contrast, we align pair-wise relationships to achieve a \textit{rotation-invariant} alignment that effectively prevents feature collapse. See Sec.~\ref {sec:motivation} for details. VGGT~\citep{vggt} is used as the 3D teacher model.}
    \label{fig:teaser}
\end{figure}
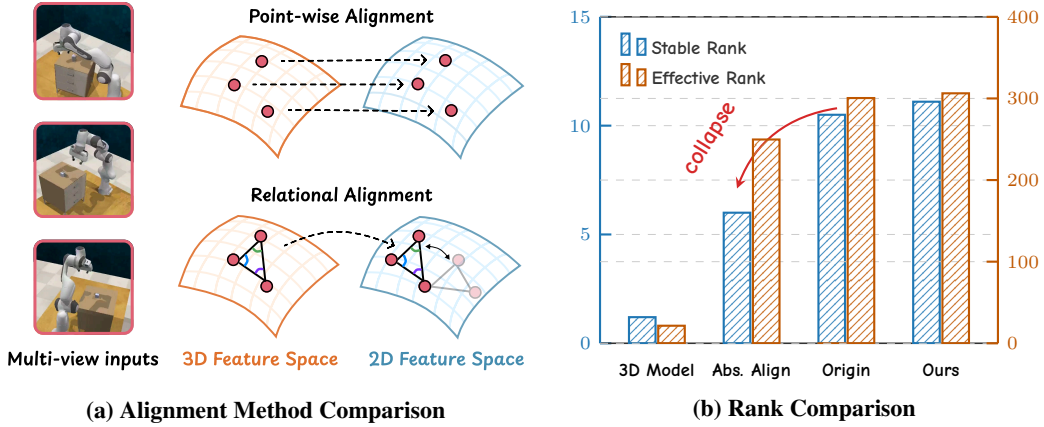

In this work, we systematically investigate the direct feature alignment between 3D foundation models and 2D VLA models. Our analysis suggests that direct spatial alignment does not necessarily lead to genuine 3D understanding. Instead, it primarily behaves as a form of low-rank regularization. Specifically, we observe a substantial collapse in the numerical rank of the VLA feature space after alignment (Fig.~\ref{fig:teaser}(b); see Sec.~\ref{sec:motivation}), indicating a significant loss of representation capacity. Consequently, experiments across diverse benchmarks show that models trained with direct spatial alignment suffer from degraded generalization performance.

To overcome these limitations, we propose a new paradigm of relational spatial alignment that aims to distill 3D awareness into 2D VLAs while preserving their expressivity. Motivated by the observation that 3D foundation models encode spatial structures through cosine similarity (i.e., the attention maps) between corresponding features, we argue that effective spatial understanding can be achieved by aligning \textit{pair-wise relationships} rather than enforcing \textit{point-wise equality}. By supervising relative similarities instead of absolute feature directions and norms, our method allows the 2D VLA representations to maintain their native feature coordinate system, with 3D knowledge being encoded in its feature similarity. (Fig.~\ref{fig:teaser}(a)). Theoretically, we demonstrate that this relational alignment objective is inherently \textbf{rotation-invariant}: The optimization objective remains strictly unchanged under arbitrary orthogonal transformations within the latent feature space, which is crucial for preventing the feature collapse. To practically instantiate this paradigm, we introduce \textbf{RIGID: Rotation-Invariant Geometry Distillation}, which implements the relation alignment paradigm and formulates the geometry knowledge transfer as an Optimal Transport problem over spatial attention distributions. Compared to direct feature supervision, RIGID improves spatial awareness while maintaining the numerical rank of the learned representations, effectively preventing capacity collapse. Extensive experiments on both simulated and real-world robotic tasks demonstrate that RIGID achieves state-of-the-art performance, exhibiting better robustness and open-world generalization than direct feature alignment.

 \section{Related work}

\boldparagraph{Explicit 3D representations in VLA models.} Recent advancements seek to enhance the spatial grounding of vision-language-action (VLA) models by directly incorporating geometric information. One main stream of methods embeds raw 3D observations, such as depth maps or point clouds, directly into the input space. Frameworks including 3D-VLA~\citep{3dvla}, 3DS-VLA~\citep{3dsvla}, PointVLA~\citep{pointvla}, 3D-CAVLA~\citep{3d-cavla}, GeoVLA~\citep{geovla}, FALCON~\cite{falcon}, and BridgeVLA~\citep{bridgevla} align these 3D inputs with 2D backbone or action experts to unify perception and manipulation, while 4D-VLA~\citep{4d-vla} extends this by stacking multiple depth maps for temporal context. Another paradigm utilizes auxiliary explicit 3D cues instead of raw sensory data. For instance, SpatialVLA~\citep{spatialvla} applies Egocentric 3D Position Encoding, and TraceVLA~\citep{tracevla} leverages visual traces. Furthermore, approaches like ReconVLA~\citep{reconvla}, DreamVLA~\citep{dreamvla}, and GuidedVLA~\citep{guidedVLA} enforce 3D representation learning by employing explicit 3D feature reconstruction as an auxiliary supervision signal. Although these models provide precision in handling height variations and object posing, they suffer from substantial data dependencies, as they often require high-fidelity 3D annotations or depth estimations that are frequently noisy in real-world environments.

\boldparagraph{Implicit 3D representations in VLA models.} Recent advances in 3D reconstruction have enabled the end-to-end learning of multi-view geometry~\cite{dust3r,mast3r,spann3r,vggt,pi3,amb3r,depthanything3}. These methods typically employ transformer-based architectures to infer geometric properties such as point maps, depth, and camera parameters from multi-view images. To circumvent the dependency on physical 3D sensors or explicit annotations, recent vision-language-action (VLA) models leverage these learned geometric cues for implicit spatial grounding. Methods such as Evo-0~\cite{evo-0, evo-1} and GeoAware-VLA~\cite{geoaware-vla} directly integrate geometric features generated by 3D foundation models into the VLA backbone. Conversely, techniques including Spatial Forcing~\cite{spatial-forcing}, ROCKET~\cite{rocket}, and GLaD~\cite{glad} leverage feature distillation. They align the intermediate latent states of the VLA with the geometric embeddings of a frozen teacher model (e.g., VGGT) to acquire spatial awareness. However, this feature integration or direct alignment across different modalities can potentially compromise the 2D image-derived representations acquired during large-scale pre-training, such as semantic understanding and general reasoning abilities, thereby degrading the overall robustness of the model.

\boldparagraph{Relational knowledge distillation.} Relational Knowledge Distillation (RKD)~\cite{rkd} is built upon the knowledge distillation (KD)~\cite{kd}, which focuses on transferring knowledge from the teacher model by distilling the in-batch structure. While traditional RKD methods primarily focus on single-modality and inter-instance associations, our approach targets the latent feature alignment between VLA models and 3D foundation models in spatial through multi-view inputs.

 \begin{figure*}
    \centering
    \includegraphics[width=1\linewidth]{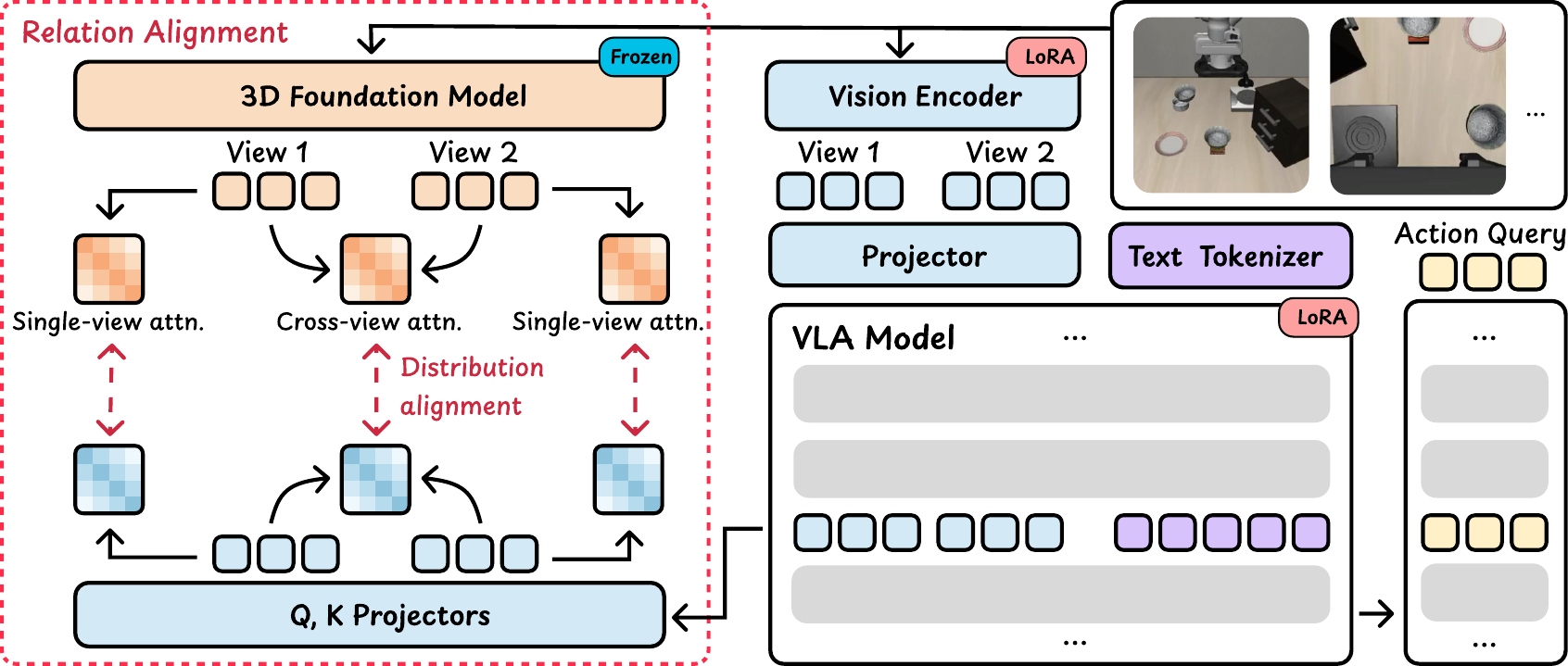}
    \caption{\textbf{Overview of our framework.} We extract self- and cross-attention (attn.) maps from both the VLA and the 3D foundation models to serve as relational affinity matrices described in Sec.~\ref{sec:motivation}. The 3D foundation model is kept frozen, and the VLA model is trained via LoRA. Other parts are fully trainable. More views can be naturally incorporated.}
    \label{fig:framework}
    \vspace{-15pt}
\end{figure*}

\section{Method}

\subsection{Motivation: feature alignment without representational collapse}
\label{sec:motivation}
Distilling 3D knowledge into 2D VLA with naive feature alignment often leads to representational collapse, as shown in Fig.~\ref{fig:teaser} (b). Specifically, let $\mathbf{Z} \in \mathbb{R}^{M \times C}$ and $\mathbf{T} \in \mathbb{R}^{M \times C_t}$ denote the feature from the student 2D VLA and the frozen 3D teacher model with sequence $M$ and channel $C$. Existing approaches~\cite{spatial-forcing, rocket} typically perform point-wise cross-domain alignment:
\begin{equation}
\mathcal{L}_{\text{existing}}(\mathbf{Z}, \mathbf{T}) = - \sum_{i=1}^{M} \frac{\mathbf{Z}_i \cdot \mathbf{T}_i}{\|\mathbf{Z}_i\| \|\mathbf{T}_i\|}.
\end{equation}
This objective forces a strict point-wise directional alignment between the two feature spaces. Consequently, the native multimodal semantic structure originally learned by the VLA model is severely distorted, leading to the rank collapse illustrated in Fig.~\ref{fig:teaser}.

Instead of enforcing point-wise alignment, we aim to preserve the relational structure between intra-domain features. Specifically, we formulate the objective using the student and teacher Gram matrices, $\mathbf{S}_{\text{vla}} = \mathbf{Z}\mathbf{Z}^\top$ and $\mathbf{S}_{\text{3D}} = \mathbf{T}\mathbf{T}^\top$:
\begin{equation}
\mathcal{L}_{\text{RIGID}}(\mathbf{Z}, \mathbf{T}) = \mathcal{D}(\mathbf{Z}\mathbf{Z}^\top, \mathbf{T}\mathbf{T}^\top),
\end{equation}
where $\mathcal{D}$ is the distance between two matrices. Unlike direct feature supervision, this objective is inherently \textbf{rotation-invariant} feature alignment: Given an arbitrary orthogonal transformation $\mathbf{R}$ satisfying $\mathbf{R}\mathbf{R}^\top = \mathbf{I}$, the optimization objective remains unchanged: 
\begin{equation}
\mathcal{L}_{\text{RIGID}}(\mathbf{Z}\mathbf{R}, \mathbf{T}) = \mathcal{D}(\mathbf{Z}\mathbf{R}\mathbf{R}^\top\mathbf{Z}^\top, \mathbf{T}\mathbf{T}^\top) = \mathcal{D}(\mathbf{Z}\mathbf{Z}^\top, \mathbf{T}\mathbf{T}^\top) = \mathcal{L}_{\text{RIGID}}(\mathbf{Z}, \mathbf{T}).
\end{equation}
This rotation invariance allows the VLA model to rotate flexibly within its own semantic coordinate basis,  thereby facilitating the preservation of its original semantic and action representations while still aligning with the geometric structure encoded by the 3D teacher. Since 3D foundation models primarily capture geometry through feature similarity~\citep{understanding_vggt}, aligning relational affinity effectively transfers spatial awareness without directly constraining original representations.

\subsection{RIGID: Rotation-invariant geometry distillation} 
\label{sec:method}

Given multi-view RGB observations $\{\mathbf{I}_t^{(v)}\}_{v=1}^{N}$, our goal is to leverage a 3D foundation model as a teacher, and distill its geometry prior into the student VLA model. 
In modern Transformers, pairwise similarity $\mathbf{Z}\mathbf{Z}^\top$ is naturally encoded in the attention responses with extra non-linearity (see Eq.~\ref{eq:attn}). Therefore, we implement $\mathcal{L}_{\text{RIGID}}$ by distilling the spatial relationships encoded within the 3D teacher model's self-attention and cross-attention maps. 
Specifically, let $\mathbf{Z}^{(v)} \in \mathbb{R}^{M \times C}$ denote the visual tokens from the VLA for the $v$-th view. For any view pair $(a,b)$, we compute the attention map of the student VLA model using learnable query and key projections $g_q(\cdot)$ and $g_k(\cdot)$:
\begin{equation}
\mathbf{A}^{(a,b)}_{\text{vla}}=
\mathrm{Softmax}
\left(
\frac{
g_q(\mathbf{Z}^{(a)})
g_k(\mathbf{Z}^{(b)})^\top
}{
\sqrt{d}
}
\right),
\label{eq:attn}
\end{equation}
The teacher attention map, $\mathbf{A}^{(a,b)}_{\text{3D}}$, is extracted directly from the frozen 3D teacher and used as a distillation target. This attention contains two complementary geometric information: \textbf{single-view attention} ($a=b$), which captures intra-view spatial layouts and local geometry, and \textbf{cross-view attention} ($a \neq b$), which models multi-view correspondences. The attention from the teacher captures rich geometric cues, providing robust spatial supervision to distill 3D awareness into the student VLA model.

\paragraph{Geometry-aware distribution alignment.} Existing distillation objectives, such as the Kullback-Leibler (KL) divergence, enforce strict token-wise correspondence. Under a KL objective, a minor spatial shift is penalized equally to a completely incorrect prediction, offering poor gradient signals when attention distributions do not perfectly overlap. Instead, we formulate $\mathcal{D}$ as an Optimal Transport (OT) problem, optimizing the discrepancy via the Wasserstein distance.

Given the student attention response $\mathbf{A}^{(a,b)}_{\mathrm{vla}}$ and the corresponding teacher attention response $\mathbf{A}^{(a,b)}_{3D}$, we interpret each row as a discrete spatial distribution over image patches. 
To measure geometric discrepancy between spatial distributions, we define a 2D transport cost matrix $\mathbf{C} \in \mathbb{R}^{M \times M}$ based on patch coordinates on the image grid. Each entry $\mathbf{C}_{p,q}$ represents the Euclidean distance between the spatial locations of the $p$-th and $q$-th patches:
\begin{equation}
\mathbf{C}_{p,q}
=
\left\|
\mathbf{c}_p - \mathbf{c}_q
\right\|_2,
\end{equation}
where $\mathbf{c}_p$ and $\mathbf{c}_q$ denote the corresponding grid coordinates.
We then compute the Wasserstein-based transport distance between the student and teacher attention distributions using entropic-regularized Optimal Transport:
\begin{equation}
\mathcal{L}_{\text{RIGID}} = \mathcal{D}(\mathbf{A}_{\text{vla}}, \mathbf{A}_{\text{3D}}) = \frac{1}{V^2 M} \sum_{a=1}^{V}\sum_{b=1}^{V} \sum_{i=1}^{M} \mathrm{Sinkhorn} \left( \mathbf{A}_{\text{vla}, i}^{(a,b)}, \mathbf{A}_{\text{3D}, i}^{(a,b)}, \mathbf{C} \right),
\end{equation}
where $V$ denotes the number of views and $M$ denotes total number of query tokens, $\mathbf{A}_{vla, i}^{(a,b)}$ and $\mathbf{A}_{3D, i}^{(a,b)}$ represent the attention distributions of the student and teacher models for the $i$-th token, respectively, and $\mathrm{Sinkhorn}(\cdot)$ computes the regularized Wasserstein distance.

The final training objective combines the action learning objective and the cross-domain geometric distillation objective:
\begin{equation}
\mathcal{L}_{\mathrm{total}}
=
\mathcal{L}_{\mathrm{action}}
+
\beta
\mathcal{L}_{\mathrm{RIGID}},
\end{equation}
where $\beta$ controls the strength of geometric supervision. By optimizing the Wasserstein distance between attention distributions, the VLA branch learns geometry-aware spatial reasoning from the 3D teacher while remaining compatible with the original transformer representation space.

 \section{Simulation}

\subsection{Setup}

\boldparagraph{Datasets and benchmarks.} We train our model on the LIBERO~\cite{libero} training set. We evaluate our proposed method across three simulated benchmarks: LIBERO, LIBERO-Plus, and our newly introduced LIBERO-Probing suite. We did not utilize the LIBEO-Plus training set.

\boldparagraph{LIBERO-Probing.} We introduce LIBERO-Probing, a new benchmark to systematically evaluate the representation capacity. Specifically, we re-render the LIBERO training environments to obtain ground-truth depth, semantic masks, and spatial correspondences. By freezing the pre-trained VLA and training a lightweight MLP, we can perform an evaluation of the learned representations using the same demonstration states as in the main task.

\boldparagraph{Implementation details.} Our training procedure on LIBERO follows OpenVLA and Spatial Forcing~\cite{openvla, openvla-oft, spatial-forcing}. The base model utilizes a VLM backbone pre-trained on the Open-X-Embodiment dataset~\cite{openx}, incorporating a fused vision encoder that combines features from SigLIP~\cite{siglip} and DINOv2~\cite{dinov2}. To ensure a fair comparison with existing baselines, we train our method using LoRA~\cite{lora}. More details can be found in the supplementary material.

\newcommand{\first}[1]{\cellcolor{blue!15}\textbf{#1}}       
\newcommand{\secon}[1]{\cellcolor{blue!5}#1}
\newcommand{\third}[1]{\cellcolor{blue!5}#1}

\begin{table}[htbp]
    \centering
    
\begin{minipage}[t]{0.432\textwidth}
        \vspace{0pt}
        \centering
        \scriptsize
        \setlength{\tabcolsep}{1pt}
        \resizebox{\linewidth}{!}{
        \begin{tabular}{lccccc}
        \toprule
          \multicolumn{1}{c}{Method}  &
          Spatial&
          Object&
          Goal &
          Long &
          Avg \\ \midrule
          \rowcolor{gray!15} \multicolumn{6}{l}{\textit{2D VLA}} \\ 
          Diffusion Policy~\cite{diffusion-policy} & 78.3 & 92.5 & 68.3 & 50.5 & 72.4 \\
          TraceVLA~\cite{tracevla} &  84.6 & 85.2 & 75.1 & 54.1 & 74.8 \\
          Octo~\cite{octo} & 78.9 & 85.7 & 84.6 & 51.1 & 75.1 \\
          Openvla~\cite{openvla} & 84.7 & 88.4 & 79.2 & 53.7 & 76.5 \\
          Dita~\cite{dita} & 84.2 & 96.3 & 85.4 & 63.8 & 82.4 \\
          CoT-VLA~\cite{cot-vla} & 87.5 & 91.6 & 87.6 & 69.0 & 83.9 \\
          $\pi_0$-FAST~\cite{pi0-fast} & 96.4 & 96.8 & 88.6 & 60.2 & 85.5 \\
          $\pi_0$~\cite{pi0} & 96.8 & 98.8 & 95.8 & 85.2 & 94.2 \\
          UniVLA~\cite{univla-rss} & 96.5 & 96.8 & 95.6 & 92.0 & 95.2 \\
          $\pi_{0.5}$~\cite{pi05} & 98.8& 98.2& 98.0& 92.4& 96.9 \\
          Openvla-OFT~\cite{openvla-oft} & 97.6 & 98.4 & 97.9 & 94.5 & 97.1 \\
\midrule
          \rowcolor{gray!15} \multicolumn{6}{l}{\textit{Explicit 3D VLA: w/ depth sensors}} \\
          SpatialVLA~\cite{spatialvla} & 88.2 & 89.9 & 78.6 & 55.5 & 78.1 \\
          4D-VLA~\cite{4d-vla} &88.9& 95.2& 90.9& 79.1& 88.6 \\
          GeoVLA~\cite{geovla} & 98.4 & 99.0 & 96.6 & \first{96.6} &97.7 \\
          3D-CAVLA~\cite{3d-cavla} & 98.2 &\first{99.8} & 98.2 & \secon{96.1} & 98.1 \\
          \midrule
          \rowcolor{gray!15} \multicolumn{6}{l}{\textit{Implicit 3D VLA: w/ RGB inputs only}} \\
          GLaD~\cite{glad} &95.2 &95.4 & 91.9 & 87.5&92.5 \\
          DreamVLA~\cite{dreamvla} &97.5 &94.0 & 89.5 & 89.5& 92.6 \\
          Spatial Forcing~\cite{spatial-forcing} & \secon{99.4} & \secon{99.6} & \first{98.8} & 96.0 & \first{98.5} \\ 
           Ours & \first{99.6} & \first{99.8} & \secon{98.4} & \secon{96.1} & \first{98.5}  \\
          \bottomrule
        \end{tabular}
        }
        \vspace{5pt}
        \caption{\textbf{Results on LIBERO.}}
    \end{minipage}\hfill
    \begin{minipage}[t]{0.54\textwidth}
        \vspace{0pt} 
        \label{tab:libero-results}
        \centering
        \scriptsize
        \setlength{\tabcolsep}{1.5pt}
        \resizebox{\linewidth}{!}{ 
        \begin{tabular}{l c c c c c c c c}
        \toprule
        Model & Cam & Ro & Lang & Light & Back & Noise &Lay &Avg \\        
        \midrule
        \rowcolor{gray!15}\multicolumn{9}{l}{\textit{LIBERO-Spatial}} \\
        \quad OpenVLA & 88.3 & 40.0 & 80.5 & 98.3 & 97.3 & 96.3 & 93.9 & 84.0 \\
        \quad Spatial Forcing & 70.2 & 43.4 & 87.1 & 99.3 & 97.7 & 88.9 & 95.1 & 82.3 \\
        \quad Ours &90.4 &51.1 &77.4 &98.6 &99.2 &96.0 &97.9 &86.6\\
        \midrule
        \rowcolor{gray!15}\multicolumn{9}{l}{\textit{LIBERO-Object}} \\
        \quad OpenVLA & 38.9 & 25.4 & 99.0 & 73.7 & 97.6 & 72.3 & 78.1 & 66.5 \\
        \quad Spatial Forcing & 49.0 & 37.9 & 99.7 & 99.3 & 98.0 & 42.2 & 72.7 & 67.7 \\
        \quad Ours & 49.0 & 37.2 & 98.3 & 97.6 & 100.0 & 82.5 & 73.2 & 74.3\\
        \midrule
        \rowcolor{gray!15}\multicolumn{9}{l}{\textit{LIBERO-Goal}} \\
        \quad OpenVLA & 62.0 & 25.2 & 53.2 & 93.9 & 92.5 & 75.2 & 59.1 & 63.0 \\
        \quad Spatial Forcing & 74.0 & 31.8 & 64.0 & 95.3 & 85.4 & 80.5 & 64.0 & 68.6 \\
        \quad Ours &74.6 &50.6 &57.8 &90.7 &88.6 &82.6 &67.3 &71.1 \\
        \midrule
        \rowcolor{gray!15}\multicolumn{9}{l}{\textit{LIBERO-Long}} \\
        \quad OpenVLA & 38.7 & 38.2 & 87.0 & 89.4 & 86.8 & 63.5 & 76.9 & 66.4 \\
        \quad Spatial Forcing & 34.1 & 35.9 & 89.9 & 78.5 & 83.7 & 55.0 & 84.3 & 63.2 \\
        \quad Ours &34.4 &50.1 &87.5 &92.7 &88.7 &55.0  & 88.1 & 67.8 \\
        \midrule
        \rowcolor{gray!15}\multicolumn{9}{l}{\textit{Total four tasks}} \\
        \quad OpenVLA-OFT & 56.4 & 31.9 & 79.5 & 88.7 & \secon{93.3} & 75.8 & 74.2 & 69.6 \\
        \quad $\pi_0$& 13.8 & 6.0 & 58.8 & 85.0 & 81.4 & \first{79.0} & 68.9 & 53.6 \\
        \quad Spatial Forcing& \secon{56.5} & \secon{37.0} & \first{84.6} & \secon{93.3} & 90.8 & 65.1 & \secon{78.3} & \secon{70.3} \\
        \quad Ours &\first{62.1} &\first{47.3} &\secon{80.3} &\first{94.9} &\first{94.1} &\first{79.0} &\first{81.6} &\first{75.0} \\  
        \bottomrule
        \end{tabular}
        }
        \vspace{5pt}
        \caption{\textbf{Results on LIBERO-Plus.}}
        \label{tab:liberoplus}
    \end{minipage}    
    
\end{table}

\subsection{Benchmark results}
In this section, we compare RIGID with state-of-the-art VLAs on simulation benchmarks on different aspects. Detailed analysis of each component, the design choices of VGGT and VLA layers, and other hyperparameters can be found in the supplementary material. 

We present a comprehensive evaluation on the LIBERO and the more challenging LIBERO-Plus benchmark. We categorize the baselines into three paradigms: (i) \textit{2D VLAs} trained on purely 2D data, (ii) \textit{Explicit 3D VLAs} requiring additional geometric modalities (e.g., depth or point clouds), and (iii) \textit{Implicit 3D VLAs} that, like ours, distill geometric knowledge in latent space, and thus do not require additional sensors at test time. 

On LIBERO, our model consistently matches or outperforms the performance of state-of-the-art implicit and explicit 3D VLAs, demonstrating that our relational alignment successfully distills geometric priors from 3D foundation models. On LIBERO plus (Tab.~\ref{tab:liberoplus}), while naive spatial alignment methods like Spatial Forcing~\citep{spatial-forcing} suffer from performance degradation under severe domain shifts due to representational collapse, our method yields superior overall robustness across diverse environmental conditions. This empirical result shows the advantage of aligning relational affinities rather than absolute feature coordinates.

\begin{figure}[t]
    
\definecolor{nblue}{HTML}{1F77B4}
    \definecolor{norange}{HTML}{BF5700}
\pgfplotsset{
        neurips_style/.style={
            width=\linewidth, height=5.5cm, 
            label style={font=\comicneue\bfseries\footnotesize},
            tick label style={font=\comicneue\bfseries\scriptsize},
            /pgf/number format/assume math mode=true,
            legend style={
                font=\comicneue\bfseries\scriptsize, 
                draw=none, 
                fill=white, 
                fill opacity=0.8, 
                cells={anchor=west},
                at={(0.5, 0.98)},      anchor=north,          legend columns=-1,     column sep=0.5ex,
                /tikz/column 2/.style={column sep=2ex}
            },
            ymajorgrids=true,
            grid style={dashed, line width=.2pt, draw=gray!40},
            axis line style={line width=.6pt, black!80},
            tick style={line width=.6pt, black!80},
        }
    }
\begin{minipage}[t]{0.43\textwidth}
        \vspace{0pt} \centering
        \includegraphics[width=\linewidth]{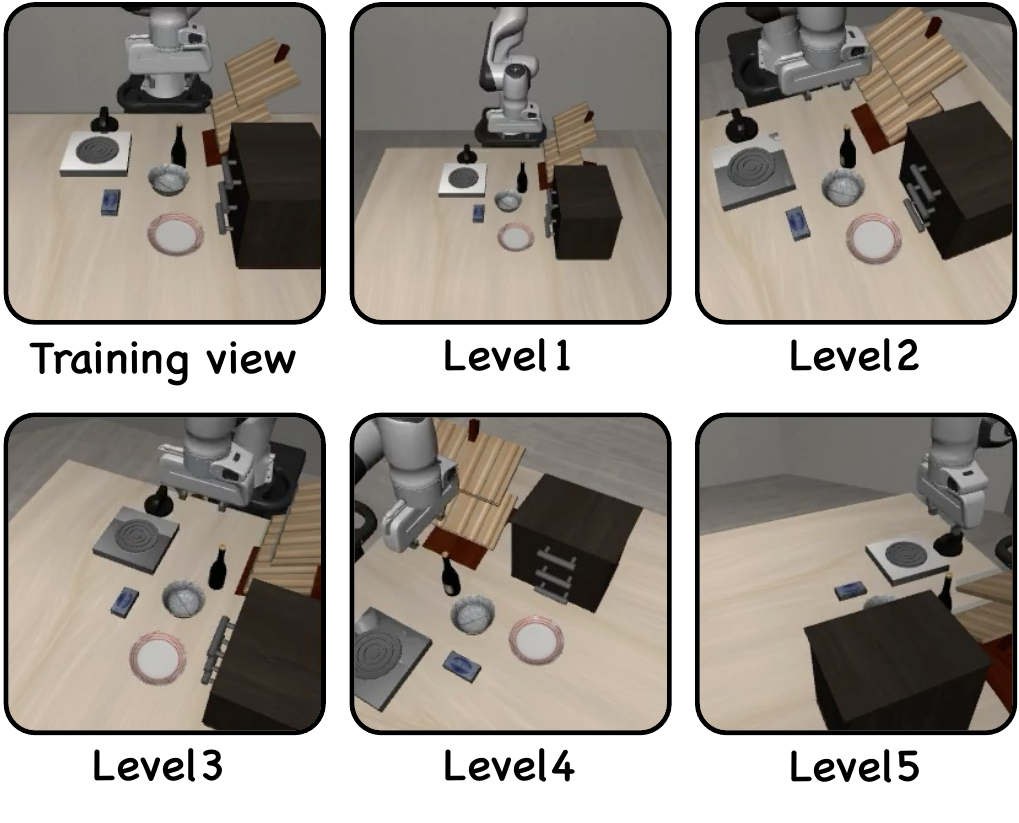}
\end{minipage}\hfill
\begin{minipage}[t]{0.55\textwidth}
    \vspace{0pt} \centering   \begin{tikzpicture}
        \pgfplotsset{
            rank_style/.style={
                neurips_style,
                enlarge x limits=0.15,
                symbolic x coords={Level 1, Level 2, Level 3, Level 4, Level 5},
                xtick=data,
                xtick style={draw=none},   nodes near coords,         every node near coord/.append style={
                    font=\comicneue\bfseries\tiny, 
                    /pgf/number format/fixed, 
                    /pgf/number format/precision=1
                },
                ybar=2pt,                  bar width=10pt,            legend image code/.code={
                    \draw[##1] (0pt,-0.5ex) rectangle (0.35cm,0.7ex);
                }
            }
        }

        \begin{axis}[
            rank_style,
            ymin=40, ymax=110,              ylabel={Success Rate (\%)},
            axis line style={rounded corners=4pt},
            y axis line style={draw=black!80, thick},
            ytick style={color=black!80, thick},
            yticklabel style={color=black},
        ]

\addplot[
                fill=norange!20, 
                draw=norange, 
                thick, 
                pattern=north west lines, pattern color=norange!80
            ] coordinates {
                (Level 1, 92.5) 
                (Level 2, 75.2) 
                (Level 3, 56.1) 
                (Level 4, 70.0) 
                (Level 5, 45.5)
            };
            \addlegendentry{Spatial Forcing}
        
\addplot[
                fill=nblue!20,            draw=nblue, 
                thick,
                pattern=north east lines, pattern color=nblue!80    ] coordinates {
                (Level 1, 94.0) 
                (Level 2, 92.5) 
                (Level 3, 92.9) 
                (Level 4, 85.0) 
                (Level 5, 63.6)
            };
            \addlegendentry{Ours}
        \end{axis}
    \end{tikzpicture}
    \end{minipage}

\caption{\textbf{Robustness to camera view changes.} We evaluate results on camera view change by level on LIBERO-Spatial. Our method outperforms the Spatial Forcing across most levels, showing particular robustness at higher difficulty tiers, mitigating the performance degradation.}
    \label{fig:cam_levels}
\end{figure}

\begin{figure*}[t]
    \centering
    
\begin{minipage}[t]{0.43\textwidth}
        \vspace{0pt} \centering
        \includegraphics[width=\linewidth]{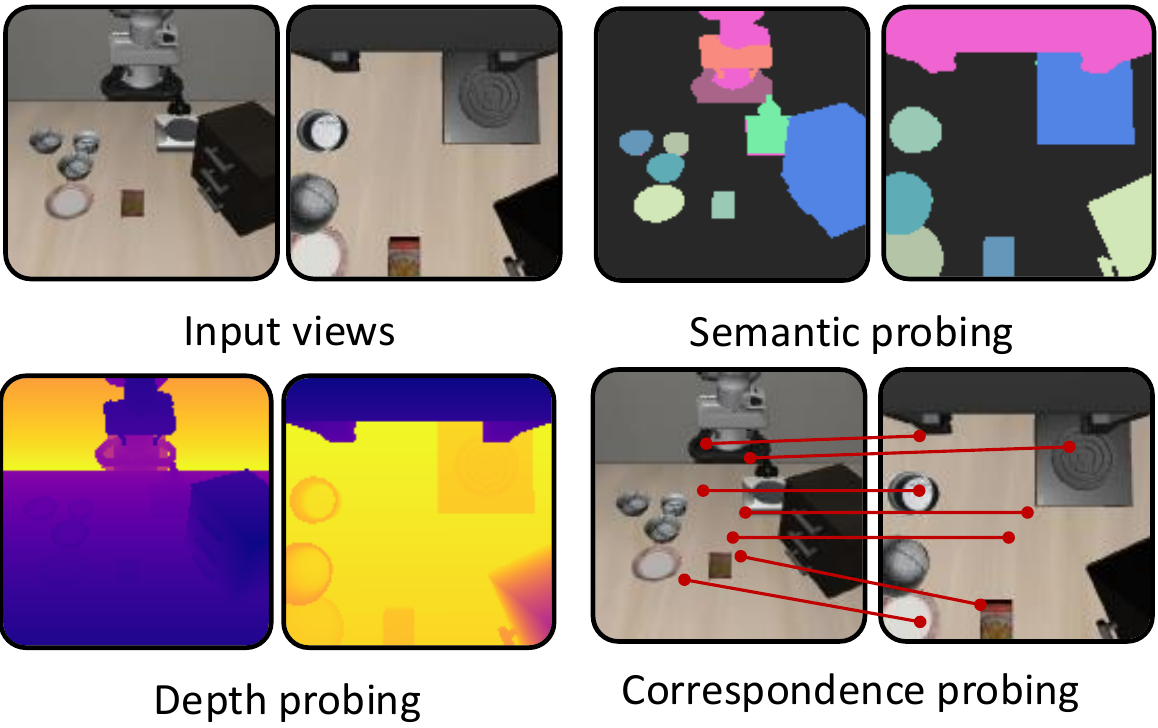}
\end{minipage}\hfill
\begin{minipage}[t]{0.54\textwidth}
        \vspace{0pt} \centering
        \small 
\resizebox{\linewidth}{!}{
        \begin{tabular}{llccc}
            \toprule
            Task & Metric & \begin{tabular}[c]{@{}c@{}}OpenVLA\\-OFT\end{tabular} & SF & Ours \\
            \midrule
            Seg. & Pixel Acc. ($\uparrow$) & 94.76 & 96.82 & \textbf{96.94} \\
            \midrule
            Depth & $\delta_{1.25}$ ($\uparrow$) & 88.12 & 92.27 & \textbf{93.60} \\
             & $\delta_{1.03}$ ($\uparrow$) & 25.30 & 36.88 & \textbf{46.90} \\
             & REL (\%) ($\downarrow$) & 10.93 & 8.05 & \textbf{6.81} \\ 
            \midrule
            Corr. & MPE ($\downarrow$) & 34.10 & \textbf{25.60} & 30.55 \\
             & PCK10 ($\uparrow$) & 0.26 & 0.29 & \textbf{0.33} \\
            \bottomrule
        \end{tabular}
        }
\end{minipage}
\caption{\textbf{Proposed LIBERO-Probing benchmark.} Our benchmark generates the semantic segmentation (Seg.), depth estimation (Depth), and correspondence matching (Corr.) results in the LIBERO training environments to analyze the VLA capabilities.}
    \label{fig:probing}
\end{figure*}

\subsection{Representation Evaluation}

While downstream task success demonstrates the effectiveness of the learned policy, it does not directly reveal whether the learned representations acquire stronger spatial and semantic understanding. We therefore evaluate the learned representations from two complementary perspectives: robustness to viewpoint changes and explicit
semantic/geometric probing.

\boldparagraph{Robustness to viewpoint changes.} To evaluate spatial understanding, we further measure robustness to viewpoint variations on LIBERO-Plus. Since these novel viewpoints are excluded from the training set and no viewpoint augmentation is applied during training, this task provides a direct evaluation of the model's 3D awareness. As shown in Fig.~\ref{fig:cam_levels}, our method consistently outperforms Spatial Forcing across all levels of viewpoint perturbation and maintains a high success rate even under the extreme viewpoint shifts (Level 4 and 5). These results demonstrate that our rotation-invariant relational alignment effectively distills robust spatial awareness into the VLA representations.

\boldparagraph{Semantic and geometric probing.} We further analyze the semantic and geometric awareness of the learned representations through a set of probing experiments. Specifically, we freeze the learned features and train lightweight MLP heads for downstream probing tasks, including depth estimation, semantic segmentation, and keypoint matching. As in Fig.~\ref{fig:probing}, our method achieves better performance across both semantic and geometric tasks. These results empirically verify that the proposed relational alignment enhances both geometric awareness and semantic understanding compared to direct feature alignment

\boldparagraph{Rank intervention.} We also examine whether rank preservation contributes to policy performance by varying spectral entropy. For point-wise alignment, encouraging higher spectral entropy increases both effective rank and success rate. For RIGID, encouraging lower spectral entropy reduces both. These results suggest that preserving rank benefits policy performance~\cite{rankme}, although we cannot attribute the performance changes to rank alone, since the regularization may also affect optimization in other ways. Experimental settings and numerical results are provided in the appendix.

\subsection{Component analysis}\label{sec:component_analysis}
We examine the distillation targets, view alignment, and teacher layer selection in RIGID. All variants are trained on the LIBERO training set for 30k steps and evaluated zero-shot on the LIBERO-Plus Spatial task suite over 2,402 episodes. Additional ablations of loss functions and other components are provided in the appendix.

\begin{figure}[tbp]
    \centering
    \includegraphics[width=\linewidth]{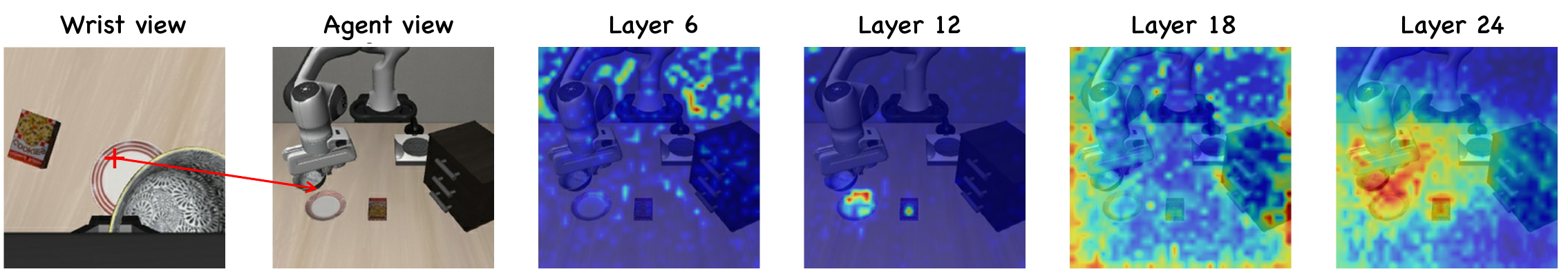}
    \caption{\textbf{Cross-view attention across VGGT layers.} The red markers indicate the query point in the wrist view and its correspondence in the agent view. Attention maps from layers (6,12,18,24) are overlaid on the agent view. }
    \label{fig:main_cross_view_attn}
\end{figure}

\newsavebox{\componenttargetbox}
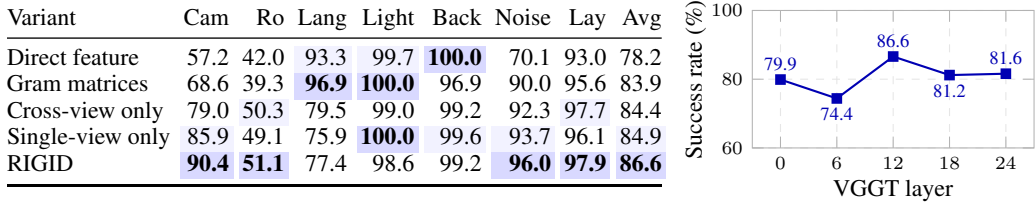
\begin{figure}[tbp]
    \centering
\begin{lrbox}{\componenttargetbox}
        \begin{minipage}[t]{0.62\linewidth}
            \vspace{0pt}
            \begingroup
    \centering
\label{tab:main_component_targets}
    \footnotesize
    \setlength{\tabcolsep}{1.5pt}
    \renewcommand{\arraystretch}{1.0}
    \begin{tabular*}{\linewidth}{@{\extracolsep{\fill}}lrrrrrrrr@{}}
        \toprule
        Variant & Cam & Ro & Lang & Light & Back & Noise & Lay & Avg \\
        \midrule
        Direct feature & 57.2 & 42.0 & \secon{93.3} & \secon{99.7} & \first{100.0} & 70.1 & 93.0 & 78.2 \\
        Gram matrices & 68.6 & 39.3 & \first{96.9} & \first{100.0} & 96.9 & 90.0 & 95.6 & 83.9 \\
        Cross-view only & 79.0 & \secon{50.3} & 79.5 & 99.0 & 99.2 & 92.3 & \secon{97.7} & 84.4 \\
        Single-view only & \secon{85.9} & 49.1 & 75.9 & \first{100.0} & \secon{99.6} & \secon{93.7} & 96.1 & \secon{84.9} \\
        RIGID & \first{90.4} & \first{51.1} & 77.4 & 98.6 & 99.2 & \first{96.0} & \first{97.9} & \first{86.6} \\
        \bottomrule
    \end{tabular*}
    \par
\endgroup
         \end{minipage}
    \end{lrbox}
    \nextfloat
    \usebox{\componenttargetbox}\hfill
    \begin{subfigure}[t]{0.38\linewidth}
    \vspace{-2pt}
    \centering
    \pgfplotsset{component layer axis/.style={
        width=\linewidth,
        height=3.4cm,
        ylabel={Success rate (\%)},
        ymin=60, ymax=100,
        ytick={60,80,100},
        grid=major,
        grid style={dashed,gray!20},
        tick label style={font=\scriptsize},
        label style={font=\footnotesize},
        axis line style={gray!60},
        major tick style={gray!60}
    }}
        \begin{tikzpicture}
            \begin{axis}[component layer axis,
                xlabel={VGGT layer},
                xlabel style={yshift=3pt},
                ylabel style={yshift=-5pt},
                xmin=-3, xmax=27,
                xtick={0,6,12,18,24}]
                \addplot[color=blue!70!black,mark=square*,thick,mark size=1.6pt]
                    coordinates {(0,79.9) (6,74.4) (12,86.6) (18,81.2) (24,81.6)};
\node[font=\scriptsize,text=blue!70!black,above] at (axis cs:0,79.9) {79.9};
                \node[font=\scriptsize,text=blue!70!black,below] at (axis cs:6,74.4) {74.4};
                \node[font=\scriptsize,text=blue!70!black,above] at (axis cs:12,86.6) {86.6};
                \node[font=\scriptsize,text=blue!70!black,below] at (axis cs:18,81.2) {81.2};
                \node[font=\scriptsize,text=blue!70!black,above] at (axis cs:24,81.6) {81.6};
            \end{axis}
        \end{tikzpicture}
        \setcounter{subfigure}{1} \phantomsubcaption
        \label{fig:main_vggt_layer}
    \end{subfigure}
    \caption{\textbf{Component analysis.} Left side: success rates (\%) on LIBERO-Plus Spatial; higher is better. Bold indicates the best result. Right side: success rates across VGGT layers. Layer 0 means the output from the VGGT encoder.}
\label{fig:main_component_layers}
\end{figure}
 
\boldparagraph{Distillation targets.}
We compare three distillation targets from VGGT: hidden features for direct point-wise alignment, Gram matrices computed as pairwise inner products of hidden features, and post-softmax query-key attention maps. The direct feature distillation variant is retrained with matched hyperparameters. As shown in Fig.~\ref{fig:main_component_layers} (left), although the Gram-matrix target achieves comparable or better performance in some perturbation categories, it performs much worse under camera perturbations (68.6\% vs.\ 90.4\%). We hypothesize that VGGT's query and key projections filter out information irrelevant to geometric correspondence, allowing attention maps to provide more effective geometric supervision. We also compare single-view and cross-view alignment in Fig.~\ref{fig:main_component_layers} (left). Combining both yields the best performance (86.6\%), suggesting that the two kinds of views provide complementary supervision.

\boldparagraph{Teacher attention and layer selection.}
We compare VGGT’s cross-view attention maps at different layers in Fig.~\ref{fig:main_cross_view_attn}. For the query point in the wrist view, layer 12 concentrates attention near the corresponding region in the agent view. Layer 6 produces scattered responses, while layers 18 and 24 show more diffuse patterns. This observation agrees with the findings of~\citet{vggt_ana} that the geometry knowledge starts to emerge in VGGT's middle layers. We also provide quantitative results for attention targets from different teacher layers in Fig.~\ref{fig:main_component_layers} (right), showing Layer 12 achieves the highest success rate among the tested layers.

 \section{Real-world experiments}

\subsection{Experimental setting}
\label{sec:setting}
\begin{figure}[t]
    \centering
    \includegraphics[width=1\linewidth]{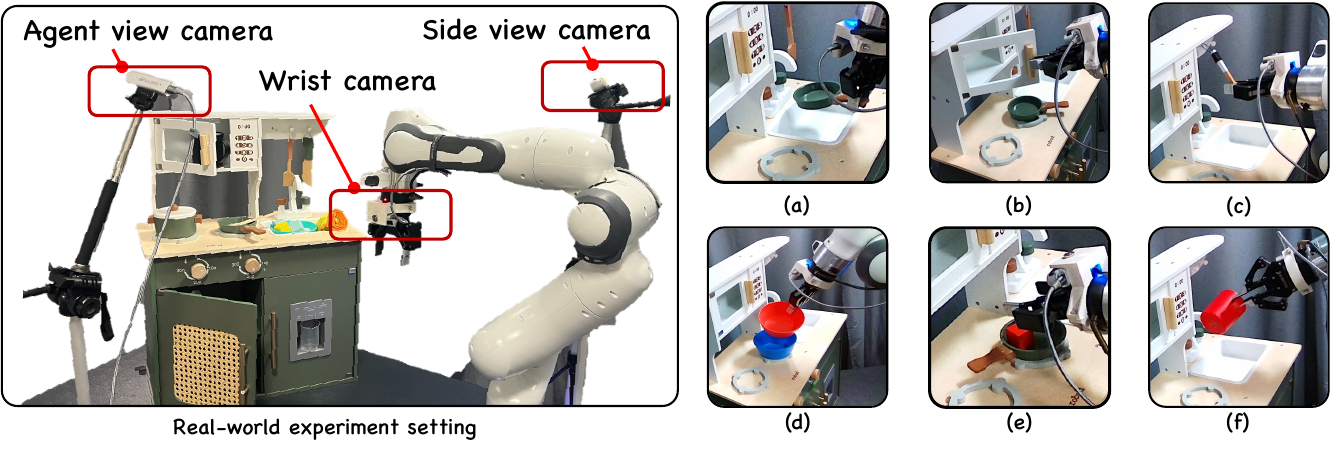}
    \vspace{-15pt}\caption{\textbf{Our robot environment and task examples.} We use a 7-DoF robotic arm with a two-finger gripper, a side-view camera, a wrist camera, and an agent-view camera (used only for visualization). We evaluate on six tasks in a complex kitchen environment.}
    \label{fig:real_robot}
\end{figure}

\boldparagraph{Setup.} We perform the real robot experiments in a complex toy kitchen environment that contains various objects. In this setting, we define six tasks as shown in Fig.~\ref{fig:real_robot}: (a) \textit{placing the pot in a sink.}, (b) \textit{opening the microwave}, (c) \textit{taking down the fork}, (d) \textit{stacking the bowls}, (e) \textit{putting the block in pan}, and (f) \textit{pouring water into the sink}. As shown in Fig.~\ref{fig:real_robot}, these scenarios require the model to handle complex environments and demonstrate spatial reasoning ability.

We employ a Franka Emika Panda robotic arm with wrist and side cameras for all tasks. An additional agent-view camera is reserved for visualization only. Both teleoperated demonstrations and robot manipulations operate at a control frequency of 15 Hz. For each task, we collect 10 demonstrations, yielding 60 task-specific trajectories. These trajectories are included in the full multitask dataset of 176 demonstrations, with the remaining demonstrations collected from other auxiliary tasks. For model training, we adopt a two-stage fine-tuning strategy: the model is first fine-tuned on the entire 176-demonstration dataset for 10k steps, followed by a task-specific fine-tuning for 1k steps using the corresponding 10 demonstrations. Finally, we evaluate the learned policies using 10 trials per task.

\boldparagraph{Baselines.} We select two representative baseline models: $\pi_0$~\citep{pi0} as our 2D VLA baseline, and Spatial Forcing~\citep{spatial-forcing} as its 3D counterpart. Following Spatial Forcing~\citep{spatial-forcing}, we use the $\pi_0$ as the backbone and integrate our proposed relation alignment. Our method and all baselines load the $\pi_0$ checkpoints pre-trained on the DROID dataset~\cite{droid} for initialization. 

\subsection{Results on real-world environment}

Fig.~\ref{fig:real_exp} shows per-task success rates across all three models. Our method achieves the best overall performance, with gains on tasks that demand precise 3D spatial reasoning. The advantage is most evident in (e)~\textit{putting the block in pan}: the policy must localize a small object from only two RGB views, and without geometric grounding, the 2D baseline $\pi_0$ fails to precisely localize the object in 3D space, while both 3D-aware models recover the correct position more reliably. For (c)~\textit{taking down the fork}, (d)~\textit{stacking the bowls}, and (f)~\textit{pouring water into the sink}, the wrist-camera view provides sufficient spatial cues for localization, so all models perform comparably well. The most challenging tasks are (a)~\textit{placing the pot in the sink} and (b)~\textit{opening the microwave}: the affordance region is small, flat, and far from the side camera, making it difficult for all the models to infer precise manipulation.

\definecolor{drawVLA}{HTML}{2CA02C}
\definecolor{fillVLA}{HTML}{2CA02C}

\definecolor{drawPI}{HTML}{BF5700}
\definecolor{fillPI}{HTML}{BF5700}

\definecolor{drawOUR}{HTML}{1F77B4}
\definecolor{fillOUR}{HTML}{1F77B4}

\begin{figure}[t]
    \centering
    \begin{tikzpicture}
        \begin{axis}[
            ybar=0pt,
            bar width=10pt,
            width=0.9\linewidth,
            height=3.5cm,
            ymin=0, ymax=100,
            ytick={0,25,50,75,100},
            ylabel={Success Rate (\%)},
            ylabel style={font=\comicneue\bfseries\footnotesize},
            tick label style={font=\comicneue\bfseries\scriptsize},
            /pgf/number format/assume math mode=true,
            ymajorgrids=true,
            grid style={dashed, gray!40},
            axis line style={black, thick},
            tick style={black, thick},
            xtick={1,2,3,4,5,6,7},
            xticklabels={(a), (b), (c), (d), (e), (f), Average},
            xticklabel style={font=\comicneue\bfseries\scriptsize},
            tick align=outside,
            enlarge x limits=0.08,
            axis x line*=bottom,
            axis y line*=left,
            legend style={
                at={(0.5,1.12)},
                anchor=south,
                legend columns=-1,
                draw=none,
                font=\comicneue\bfseries\scriptsize,
                /tikz/every even column/.append style={column sep=0.6cm}
            },
            legend image code/.code={
                \draw[#1] (0pt,-0.6ex) rectangle (0.18cm,0.8ex);
            }
        ]

        \addplot[
            fill=fillVLA,
            draw=drawVLA,
            line width=0.6pt,
            pattern=north east lines,
            pattern color=drawVLA
        ] coordinates {
           (1,10) (2, 20)  (3, 90) (4, 30)  (5, 20)  (6, 80)  (7, 41.7)
        };

        \addplot[
            fill=fillPI,
            draw=drawPI,
            line width=0.6pt,
            pattern=north east lines,
            pattern color=drawPI
        ] coordinates {
            (1,10)  (2, 40)  (3, 80)  (4, 50)  (5, 70) (6, 80) (7, 55.0)
        };

        \addplot[
            fill=fillOUR,
            draw=drawOUR,
            line width=0.6pt,
            pattern=north east lines,
            pattern color=drawOUR
        ] coordinates {
            (1,10) (2, 40)  (3, 90)  (4, 60) (5, 90)  (6, 90) (7, 63.3)
        };
        
        \legend{$\pi_{0}$, Spatial Forcing, Ours}
        
        \draw[dashed, black!60, line width=0.8pt] (axis cs:6.5, 0) -- (axis cs:6.5, 100);
        
        \end{axis}
    \end{tikzpicture}
    \caption{\textbf{Performance comparison on real-world tasks.} We report the success rates and the average rates on the six real robot tasks. All the models took 10 trials on each task. (a) – (f) are consistent with the labels described in Fig.~\ref{fig:real_robot} and Sec.~\ref{sec:setting}.}
    \label{fig:real_exp}
\end{figure}
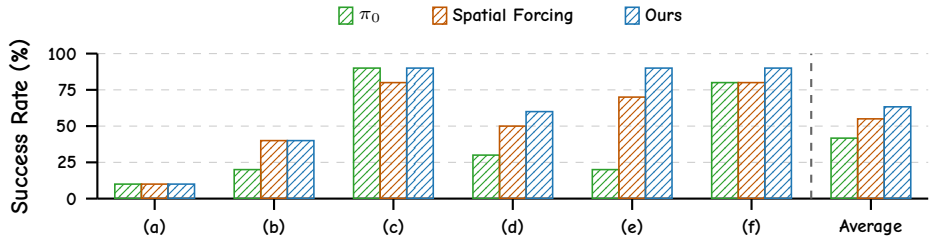
 
 \section{Discussion}

\paragraph{Limitations.} 
Despite its effectiveness, RIGID relies on a pre-trained 3D expert during training. Therefore, the quality of spatial distillation inherently depends on both the teacher model's capabilities and the scale of the training data. Specifically, we use VGGT~\cite{vggt} as the 3D teacher; while effective and generalizable, it still possesses known biases and limitations~\citep{amb3r, vggt_ana}. Furthermore, the limited scale of the robot fine-tuning dataset presents another constraint. Although VGGT is pre-trained on large-scale 3D data, the comparatively much smaller fine-tuning dataset restricts the diversity and generalizability of the spatial knowledge transferred to the student model.

\paragraph{Conclusion.} We present RIGID, a framework designed to enhance spatial understanding in VLA models while remaining effective on VLA's original capabilities. The key idea is that, unlike direct feature alignment with a 3D teacher model, RIGID distills 3D awareness into 2D VLAs while preserving their original representation space. To effectively incorporate spatial cues, we formulate geometry distillation as a Wasserstein-based optimization problem. Experiments in simulation, real-world robotic settings, and probing analyses, together with representation rank analysis, demonstrate that our method improves spatial understanding while preserving semantic understanding.

\acknowledgments{This work was supported by the Royal Society Research Grant RGS/R2/242051, and utilised the Isambard-AI National AI Research Resource (AIRR)~\cite{isambard}. }

\bibliography{main}  

\clearpage

\renewcommand{\thefigure}{S\arabic{figure}}
\renewcommand{\thetable}{S\arabic{table}}
\setcounter{figure}{0}
\setcounter{table}{0}
\appendix

\clearpage

\section{Training details}
Table~\ref{tab:combined_hyperparams} shows the default settings used in our LIBERO and real-world experiments. Other settings follow the default fine-tuning configurations for OpenVLA-OFT~\cite{openvla-oft} or $\pi_0$~\cite{pi0} in their original repositories.
\begin{table*}[h]
\centering
\small
\setlength{\tabcolsep}{10pt}        \renewcommand{\arraystretch}{1.15}  \begin{tabular}{l|l|l}
\hline
\textbf{Hyperparameters} & \textbf{Simulation (LIBERO) Config} & \textbf{Real-Robot Config} \\ \hline
Base VLA & OpenVLA (7B) & $\pi_0$-DROID (3B) \\
Optimizer & AdamW ($\beta_1{=}0.9,\ \beta_2{=}0.95$) & AdamW ($\beta_1{=}0.9,\ \beta_2{=}0.95$) \\
Learning rate & 5e-4 / 5e-5 (after 20k steps) & 2.5e-5 (peak) \\
Lr Scheduler & MultiStepLR & Cosine Decay \\
Warmup Steps & 1,000 & 1,000 \\
Batch Size & 16 & 16 \\
Training Steps & 30,000 & 10,000 \\
EMA & 0.999 & None \\
$\text{distill loss coefficient}$ & 0.5 & 0.5 \\
VLA align layers & 24 (total 32 layers)  & 12 (total 18 layers) \\
VGGT extract layers & 12  & 12  \\
Fine-tuning / LoRA rank & LoRA (rank 32) & Full (no LoRA) \\
GPUs & NVIDIA H100 80G $\times$ 2 & NVIDIA H100 80G $\times$ 1 \\ \bottomrule
\end{tabular}
\caption{Combined hyperparameters for simulation and real-robot training configurations.}
\label{tab:combined_hyperparams}
\end{table*} 
\section{Additional analysis on rank collapse}

\boldparagraph{Implementing details.}
In this section we present the implementation details for our rank analysis. We compare the visual features of our model against three baselines: OpenVLA~\cite{openvla-oft}, Spatial Forcing~\cite{spatial-forcing}, and the 3D foundation model (VGGT)~\cite{vggt}. For each model, we process 50 batches ($50 \times 8 = 400$ samples) of LIBERO-Spatial trajectories from the training split. We extract the hidden states from layer 24, where both our method and Spatial Forcing incorporate depth information, and retain only visual patch tokens. For VGGT, we directly use the representations from its final layer, which also serve as the distillation target in Spatial Forcing.

We then mean-pool the spatial patch tokens to obtain a single global feature vector per frame. After applying per-sample $\ell_2$ normalization, we construct a feature matrix and mean-center it across samples by subtracting the feature mean from each vector before performing singular value decomposition (SVD). From the resulting singular-value spectrum, we compute three metrics to assess feature diversity: the stable rank, which reflects the number of effectively utilized dimensions; the effective rank, which reflects the uniformity of variance across dimensions; and the PCA cumulative explained variance, which indicates the degree of feature collapse into dominant dimensions.

\begin{figure}
    \centering
    \begin{minipage}[t]{0.49\textwidth}
        \includegraphics[width=1.0\linewidth]{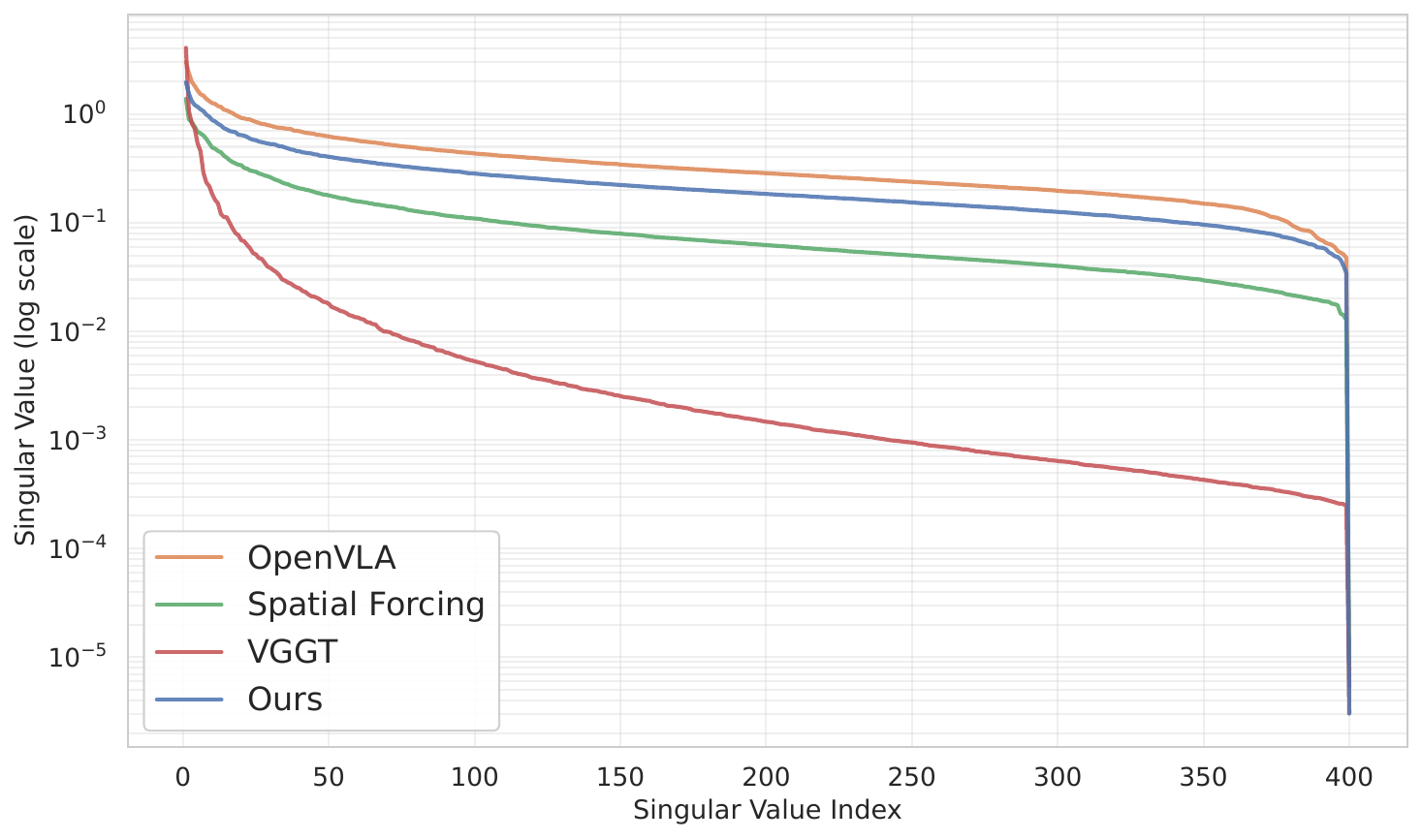}
        \subcaption{Scree Plot}
    \end{minipage}
    \hfill
    \begin{minipage}[t]{0.49\textwidth}
         \includegraphics[width=1.0\linewidth]{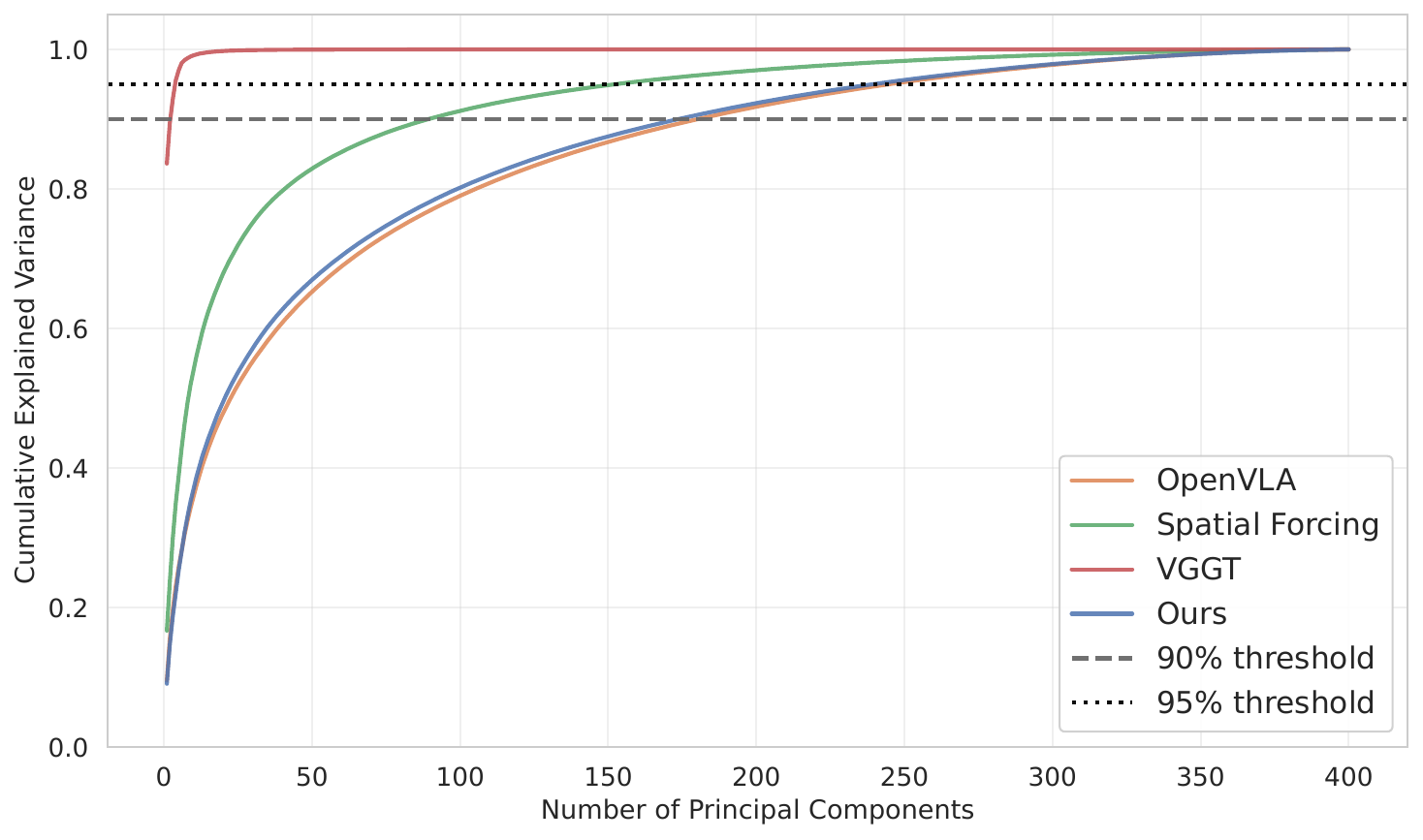}
         \subcaption{Cumulative Explained Variance Plot}
    \end{minipage}
    \caption{Dimensionality analysis of visual features.}
    \label{fig:sup_collapse}
\end{figure}

We also visualize these properties using a Scree Plot (Fig.~\ref{fig:sup_collapse}(a)) and a Cumulative Variance Plot (Fig.~\ref{fig:sup_collapse}(b)). While the absolute rank values may be affected by the limited sample size, our analysis focuses on the relative differences across models.

\boldparagraph{Additional analysis.}
In the Scree Plot (Fig.~\ref{fig:sup_collapse}(a)), the log-scaled singular values across component indices reflect the representational capacity of the model, where a slowly decaying curve indicates a rich, high-dimensional feature space. As shown, the curve for our method closely aligns with that of the OpenVLA baseline. This similarity suggests that our method preserves feature diversity relative to the OpenVLA baseline. Conversely, the curve for the Spatial Forcing baseline shows a noticeable downward shift, indicating a loss of expressive power across higher-order components. This dimensional collapse is further validated by the Cumulative Explained Variance plot (Fig.~\ref{fig:sup_collapse}(b)), where a steeper ascent implies that information is heavily compressed into fewer principal components. The Spatial Forcing baseline reaches the 90\% variance threshold significantly earlier than the VLA baseline, demonstrating that direct feature forcing disrupts the model's natural feature distribution and aggressively collapses the representation into a lower-dimensional subspace.

\boldparagraph{Rank intervention study.}
We provide evidence that representation rank contributes to policy performance via a bidirectional intervention on LIBERO-Plus Spatial (30k steps; Tab.~\ref{tab:alignment_rank}). Within each pair, we fix the alignment objective and training setup and add rank regularization.

(i) Preventing rank collapse.
For point-wise alignment, we add
\[
    \mathcal{L}_{\mathrm{rank}} = \left[\max\left(0, H_{\mathrm{pre}}-H(Z)\right)\right]^2,
\]
where $H(Z)$ is the student features' spectral entropy and $H_{\mathrm{pre}}$ is the pre-trained VLA's spectral entropy. Penalizing spectral entropy below $H_{\mathrm{pre}}$ raises both rank and performance.

(ii) Inducing rank collapse.
For RIGID, we keep $\mathcal{L}_{\mathrm{RIGID}}$ unchanged and add
\[
    \mathcal{L}_{\mathrm{collapse}} = \gamma\left(H(Z)-H^{*}\right)^2
\]
to push rank down to a lower target entropy $H^{*}$; performance drops accordingly. Here, $\gamma$ controls the regularization strength.

Restoring rank helps point-wise alignment and suppressing it hurts RIGID without touching the alignment loss, though rank is not the sole determining factor. The regularization may also affect optimization in other ways.

\definecolor{nipsred}{RGB}{180,60,60}
\definecolor{nipsgreen}{RGB}{60,140,90}

\begin{table*}[tbp]
    \centering
    \small
    \setlength{\tabcolsep}{6pt}
    \begin{tabular}{llll}
        \toprule
        Method & Alignment & Eff. Rank & Perf. (\%) \\
        \midrule
        Baseline
            & Point-wise & 248.0 & 72.9 \\
        \quad \textcolor{nipsgreen}{+ preventing rank collapse}
            & Point-wise & 258.8 \textcolor{nipsgreen}{$\uparrow$} & 78.6 \textcolor{nipsgreen}{$\uparrow$} \\
        \midrule
        Ours
            & Relational & 306.3 & \textbf{85.2} \\
        \quad \textcolor{nipsred}{+ inducing rank collapse}
            & Relational & 278.5 \textcolor{nipsred}{$\downarrow$} & 80.8 \textcolor{nipsred}{$\downarrow$} \\
        \bottomrule
    \end{tabular}
    \caption{\textbf{Rank intervention on LIBERO-Plus Spatial.} We report effective rank and policy success rate after 30k training steps.}
    \label{tab:alignment_rank}
\end{table*}

\section{Additional ablation study}
We provide additional ablation study in this section, including different losses and other design choices, under the same training and evaluation protocol as the Section~\ref{sec:component_analysis} in the main text.

\boldparagraph{VLA layers.} We conduct an ablation study on the selection of the VLA integration layer. Consistent with Spatial Forcing~\cite{spatial-forcing}, we empirically find that integrating spatial features at Layer 24 yields optimal performance. However, we observe only marginal performance variations across the different layers.

\begin{table*}[t]
    \centering
    \scriptsize
    \setlength{\tabcolsep}{1.5pt}
    \resizebox{\linewidth}{!}{ 
    \begin{tabular}{l ccccccccccccc}
    \toprule
    Model & Cam & L1 & L2 & L3 & L4 & L5 & Ro & Lang & Light & Back & Noise & Lay & Avg \\       
    \midrule
    \rowcolor{gray!15}\multicolumn{14}{l}{\textit{Loss}} \\
    \quad MSE   & 60.1 & 73.1 & 66.9 & 53.5 & 55.0 & 22.7 & 38.0 & \first{99.0} & \first{100.0} & \first{99.6} & 75.8 & 84.4 & 78.5 \\
    \quad KL    & 73.9 & 83.6 & 76.7 & 69.3 & \secon{70.0} & 59.1 & \first{52.9} & 88.7 & 97.9  & 98.1 & 88.6 & 95.6 & 84.4 \\
    \quad L1    & 68.6 & \secon{95.5} & 79.7 & 55.3 & 42.5 & 36.4 & \secon{52.0} & 89.5 & \first{100.0} & \secon{99.2} & 81.5 & 94.5 & 82.7 \\
    \quad MMD   & 60.9 & 85.1 & 74.4 & 51.8 & 30.0 & 9.1  & 47.7 & 97.2 & \secon{99.7}  & 98.4 & 78.1 & 91.4 & 81.0 \\
    \quad Wasserstein & \secon{83.2} & \first{97.0} & \secon{88.0} & \secon{74.6} & 67.5 & \first{86.4} & 48.1 & 83.6 & \secon{99.7}  & 94.2 & \secon{90.3} & \secon{96.1} & \secon{85.2} \\
    \midrule
    \rowcolor{gray!15}\multicolumn{14}{l}{\textit{Other Ablations (based on Wasserstein loss)}} \\
    \quad (a) w/ EMA             & \first{90.4} & 94.0 & \first{92.5} & \first{92.9} & \first{85.0} & \secon{63.6} & 51.1 & 77.4 & 98.6 & \secon{99.2} & \first{96.0} & \first{97.9} & \first{86.6} \\
    \quad (b) w/ distill coe.=2.0          & 47.1 & 70.1 & 55.6 & 32.5 & 32.5 & 27.3 & 44.0 & 96.2 & \secon{99.7} & 97.3 & 68.1 & 90.6 & 76.4 \\
    \quad (c) w/ attention heads & 54.8 & 91.0 & 69.2 & 26.3 & 37.5 & 36.4 & 42.3 & \secon{98.2} & 99.3 & \secon{99.2} & 77.5 & 94.0 & 79.8 \\
    \bottomrule
    \end{tabular}
    }
    \caption{\textbf{Component analysis.} We systematically analyze different component choices. KEY-- EMA: Exponential Moving Average, coe.: coefficient.}
    \label{tab:component}
\end{table*}

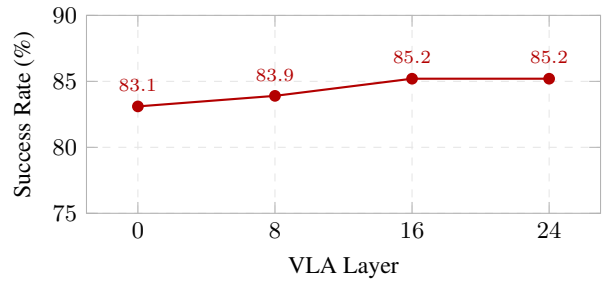
\begin{figure}[t]
    \centering
    \begin{tikzpicture}
        \begin{axis}[
            width=0.6\textwidth,
            height=4.2cm,
            xlabel={VLA Layer},
            ylabel={Success Rate (\%)},
            grid=major,
            grid style={dashed, gray!20},
            ymin=75, ymax=90,
            xmin=-3, xmax=27,
            xtick={0, 8, 16, 24},
            ytick={75, 80, 85, 90},
            tick label style={font=\footnotesize},
            label style={font=\footnotesize},
            axis line style={gray!60},
            major tick style={gray!60}
        ]
            \addplot[
                color=red!70!black,
                mark=*,
                thick,
                mark size=1.8pt,
                nodes near coords,
                nodes near coords style={
                    font=\scriptsize,
                    yshift=2pt,
                    /pgf/number format/.cd, fixed, fixed zerofill, precision=1
                }
            ] coordinates {
                (0, 83.1) (8, 83.9) (16, 85.2) (24, 85.2)
            };
        \end{axis}
    \end{tikzpicture}
    \caption{Success rate across VLA integration layers. Layer 0 denotes the output of the VLA vision encoders.}
    \label{fig:vla_layer}
\end{figure}
 
\boldparagraph{Loss function.}
We evaluated several loss functions as optimization objectives, including L1, Mean Squared Error (MSE), Kullback-Leibler (KL) divergence, Maximum Mean Discrepancy (MMD)~\citep{mmd1,mmd2}, and Wasserstein loss. Empirically, the Wasserstein loss achieves the best balance across all perturbation categories. While KL divergence performs well generally, it yields suboptimal results on camera view change tasks. This aligns with our previous analysis~\ref{sec:method}: under a KL objective, a minor spatial shift is penalized equally to a completely incorrect prediction, which provides poor gradient signals when attention distributions do not perfectly overlap. Consequently, its generalization to viewpoint variations is less stable than that of the Wasserstein loss. 

For other losses, MMD utilizes a kernel-based metric measuring the distance between probability distributions and is position-agnostic. As a result, it shows suboptimal performance on spatial understanding tasks that require precise location information. By comparison, L1 and MSE are designed to match absolute values rather than measure distribution distances. They performed poorly on spatial understanding tasks but demonstrated excellent robustness against position-invariant perturbations, such as lighting and background changes. 

Considering all these factors, we select the Wasserstein loss as the primary optimization objective for our method. However, we also observe its limitations in handling language and robot state perturbations. Effectively combining the strengths of different loss functions remains a valuable direction for future research.

\begin{samepage}
\boldparagraph{Computational cost for Wasserstein loss.}
We also measure the extra computational cost of using the Wasserstein loss. Tab.~\ref{tab:speed} reports wall-clock time, FLOPs, and peak memory vs.\ KL divergence which is commonly used for distillaiton loss (batch size 1, single H100). We find that Sinkhorn-OT only adds minimal additional computational overhead, with an increase of only 8.0\% training time and 0.09\% FLOPs relative to attention-KL, while the reported peak memory remains the same. At inference, the spatial module is dropped entirely, matching the OpenVLA baseline.
\par
\end{samepage}

\begin{table*}[tbp]
    \centering
    \caption{\textbf{Computational cost of attention-map distillation.} Measurements use a batch size of 1 on a single H100 GPU. $T$ denotes the number of Sinkhorn iterations. Percentages are relative to the KL baseline.}
    \label{tab:speed}
    \resizebox{0.8\linewidth}{!}{\begin{tabular}{lccc}
        \toprule
        Method & Train time (s/step) & FLOPs/step & Peak mem. \\
        \midrule
        + affinity distillation, KL
            & 0.6312 & 32.91 T & 27.05 GB \\
        + Sinkhorn-OT, $T=10$
            & 0.6501 (+3.0\%) & 32.92 T (+0.03\%) & 27.05 GB \\
        + Sinkhorn-OT, $T=20$
            & 0.6580 (+4.2\%) & 32.92 T (+0.03\%) & 27.05 GB \\
        + Sinkhorn-OT, $T=50$ (ours)
            & 0.6814 (+8.0\%) & 32.94 T (+0.09\%) & 27.05 GB \\
        \bottomrule
    \end{tabular}}
\end{table*}
 
\boldparagraph{EMA.} We found that the usage of Exponential Moving Average (EMA) helps stabilize the training and further improve the performance. We apply an exponential moving average over all trainable parameters (LoRA adapters + action head + alignment projectors) and evaluate with the averaged weights. We hypothesize that its benefit is attributed to two factors: the small batch size imposed by our compute budget, which makes per-step gradients noisy, and the competition among the action loss and the auxiliary alignment losses in early training; averaging the weight trajectory smooths both effects. We also perform an additional study on different EMA decay rates, as illustrated in Tab.~\ref{tab:decay_rate}. We also find that a high decay rate ($\beta=0.9999$) yields the lowest success rate (82.1\%), as overly slow EMA updates fail to track the active model, leading to evaluation on outdated parameters and reduced performance.

\begin{table*}[h]
    \centering
    \small
    \caption{Performance according to the decay parameter $\beta$ on LIBERO-Plus Spatial.
    }\label{tab:momentum}
    \setlength{\tabcolsep}{4pt}
        \begin{tabular}{lccccc}
        \toprule
         EMA decay rate   &  0.0    &   0.9    &   0.999   &   0.9999 \\
        \midrule
        Avg. successful rate (\%)    &   85.2   &  85.1    &   86.6   &   82.1    \\
        \bottomrule
        \end{tabular}
    \label{tab:decay_rate}
    \end{table*}
 
\boldparagraph{Distillation coefficient.} We set the distillation coefficient to 0.5 to prevent it from overwhelming the action supervision signal. As shown in Tab.~\ref{tab:component}(b), weighing the distillation supervision more heavily than the action signal leads to suboptimal performance.

\boldparagraph{Attention heads.} The original VGGT layers contain multiple attention heads. In our current design, we average the outputs across all heads. We also investigated incorporating a multi-head design into the projectors for the VLA vision features, performing distillation on a per-head basis. However, in practice, we found that this approach complicates training with a limited fine-tuning dataset and ultimately leads to suboptimal performance.

\section{Probing experiments}

\boldparagraph{Probing dataset construction.} We generate a dedicated probing dataset using the LIBERO benchmark~\cite{libero}, which provides 130 tabletop manipulation tasks across five task suites: LIBERO-Spatial (10 tasks), LIBERO-Object (10 tasks), LIBERO-Goal (10 tasks), LIBERO-10 (10 tasks), and LIBERO-90 (90 tasks). All demonstrations are rendered in the MuJoCo physics simulator via Robosuite~\cite{robosuite}. For each task demonstration, we render two camera views simultaneously: an agent view (fixed third-person camera) and a wrist camera. For each view we record:
\begin{itemize}
    \item RGB images from two views.
    \item Depth map in meters, from the simulation depth buffer.
    \item Instance segmentation map of per-pixel instance IDs from the simulation scene graph. Robot body components and backgrounds are excluded and remapped to class 0. The resulting taxonomy contains 52 classes: background and 51 interactive object categories. 
    \item Cross-View Correspondence. For each demonstration frame, we compute up to 32 valid point correspondences between agent-view and wrist-view using geometric reprojection through camera intrinsics and extrinsics provided by the simulation. Candidate query pixels are sampled uniformly from foreground pixels in the wrist-view image.
\end{itemize}

Each LIBERO task provides 50 demonstrations that often exhibit near-identical initial object configurations, introducing significant redundancy. Therefore, for each task, we select five maximally diverse demonstrations using farthest-point sampling (FPS) on the MuJoCo state vector. FPS initializes from demo 0 and iteratively selects the demo whose state vector is farthest (in $\ell_2$ distance) from all already-selected demos. 

We split at the task level rather than the demo level to test generalisation to held-out tasks and object configurations. For the four main suites, we assign the 8 tasks to training and the 2 tasks to testing. The LIBERO-90 suite (90 tasks) is assigned entirely to training following the setup in the original LIBERO dataset. As a result, it contains 122 training tasks of 610 demos and 8 testing tasks with 40 demos.

\boldparagraph{Feature extraction.} We extract the hidden state features from the 24th layer of OpenVLA~\cite{openvla}, Spatial Forcing~\cite{spatial-forcing}, and our proposed method. This specific layer is selected because the distillation loss is applied to it during training, thus directly reflecting how distillation shapes the learned representations. All probes operate on these frozen features using lightweight output heads, detailed as follows.

\boldparagraph{Depth probe.} The output head architecture utilizes a $1 \times 1$ convolution to map the features to spatial scalars, followed by a Softplus activation to ensure positivity and then bilinear upsampling to input resolution. The L1 loss is applied. The evaluation metrics are $\delta_{1.25}$, $\delta_{1.03}$, and Relative Error (REL).

\boldparagraph{Segmentation probe.} A $1 \times 1$ convolution projects the features into 52 class logits per spatial location, which are subsequently bilinearly upsampled. This probe is supervised via the cross-entropy loss. The evaluation metric is Pixel Accuracy.

\boldparagraph{Correspondence probe.} A shared linear projection embeds both agent and wrist view features into a 256-dimensional space, generating a spatial correspondence heatmap via $\ell_2$-normalized dot-product attention. The objective is formulated as a 256-way cross-entropy classification loss applied over the flattened $16 \times 16$ token grid. The evaluation metrics are Mean Pixel Error (MPE) and PCK10.

\section{Real robot experiments} 
\begin{figure*}[b]
    \centering
    \includegraphics[width=1\linewidth]{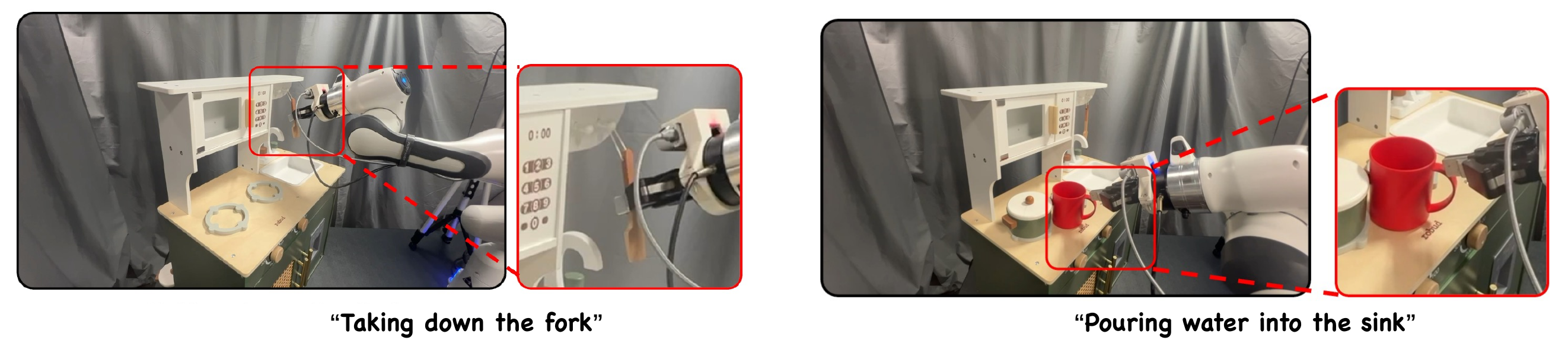}
    \caption{Failure cases}
    \label{fig:failure_cases}
\end{figure*}

\begin{figure*}[t]
    \centering
    \includegraphics[width=1\linewidth]{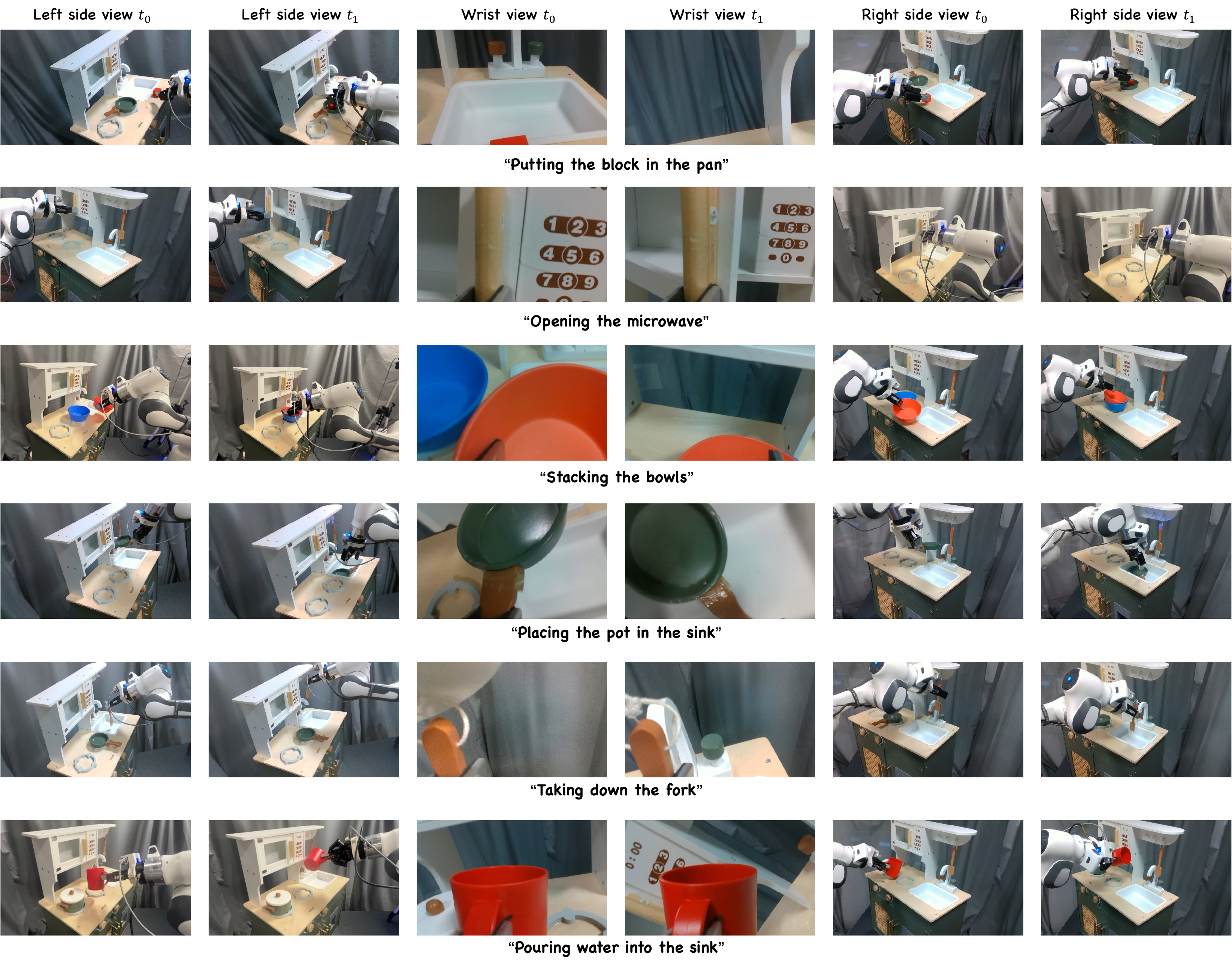}
    \caption{Real robot experiments}
    \label{fig:real_demo}
\end{figure*}

For demonstrations of the real-world robot experiments, we strongly encourage readers to review our supplementary videos. Additionally, Fig.~\ref{fig:real_demo} illustrates these experiments, detailing the six evaluation tasks and the corresponding multi-view input images. Note that the left-side view (agent view) is provided solely for visualization purposes. Finally, we report typical failure modes, such as imprecise localization and physical execution errors, as shown in~\ref{fig:failure_cases}. Please refer to the supplementary video for results and failure cases.

\end{document}